\documentclass[lettersize,journal]{IEEEtran}
\usepackage{amsmath,amsfonts}
\usepackage{booktabs} 
\usepackage{algorithmic}
\usepackage{algorithm}
\usepackage{array}
\usepackage[caption=false,font=normalsize,labelfont=sf,textfont=sf]{subfig}
\usepackage{textcomp}
\usepackage{stfloats}
\usepackage{url}
\usepackage{verbatim}
\usepackage{graphicx}
\usepackage{cite}
\usepackage{multirow}
\usepackage[inline]{enumitem}
\begin{document}

\title{Hierarchical, Interpretable, Label-Free Concept Bottleneck Model}

\author{Haodong Xie, Yujun Cai, Rahul Singh Maharjan, Yiwei Wang, Federico Tavella, and Angelo Cangelosi.
\thanks{Haodong Xie and Rahul Singh Maharjan, Federico Tavella, and Angelo Cangelosi are with the Department of Computer Science, University of Manchester, United Kingdom. (email: Haodong.Xie@manchester.ac.uk).}
\thanks{Yujun Cai is with the School of Electrical Engineering and Computer Science, the University of Queensland, Australia.}
\thanks{Yiwei Wang is with Department of Computer Science, University of California at Merced, United State.}
}



\maketitle

\begin{abstract}
Concept Bottleneck Models (CBMs) introduce interpretability to black-box deep learning models by predicting labels through human-understandable concepts. However, unlike humans, who identify objects at different levels of abstraction using both general and specific features, existing CBMs operate at a single semantic level in both concept and label space. We propose HIL-CBM, a Hierarchical Interpretable Label-Free Concept Bottleneck Model that extends CBMs into a hierarchical framework to enhance interpretability by more closely mirroring the human cognitive process. HIL-CBM enables classification and explanation across multiple semantic levels without requiring relational concept annotations. HIL-CBM aligns the abstraction level of concept-based explanations with that of model predictions, progressing from abstract to concrete. This is achieved by (i) introducing a gradient-based visual consistency loss that encourages abstraction layers to focus on similar spatial regions, and (ii) training dual classification heads, each operating on feature concepts at different abstraction levels. Experiments on benchmark datasets demonstrate that HIL-CBM outperforms state-of-the-art sparse CBMs in classification accuracy. Human evaluations further show that HIL-CBM provides more interpretable and accurate explanations, while maintaining a hierarchical and label-free approach to feature concepts.
\end{abstract}


\section{Introduction}

Taxonomy organize objects into hierarchical categories based on shared characteristics, ranging from general to specific. In the visual domain, taxonomic structures are natural and central to human understanding. Among these levels, the \textit{basic level of abstraction} holds particular significance in human categorization~\cite{rosch1976basic}. Within a taxonomic hierarchy, the basic level occupies the middle tier, representing a moderate level of generality~\cite{wang2018categorizing}. Typically, when people name objects, they first identify them at the basic level (e.g., \textit{dog}, \textit{cat})~\cite{rosch1976basic}. Categorization at the superordinate level (e.g., \textit{animal}) involves retrieving more abstract semantic knowledge, whereas subordinate-level categorization (e.g., \textit{border collie}, \textit{devon rex}) demands finer-grained perceptual analysis~\cite{jolicoeur1984pictures}.

\begin{figure}[t]
\centering
\includegraphics[width=0.48\textwidth]{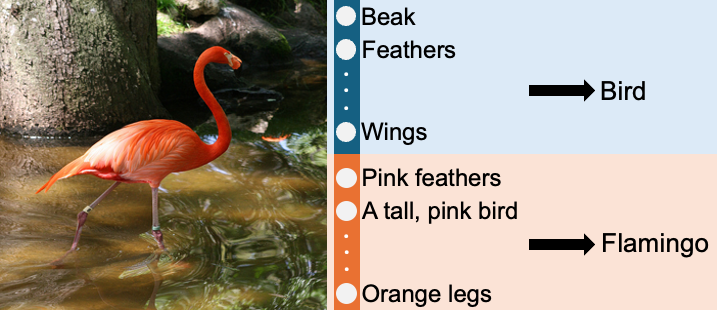} 
\caption{Our proposed HIL-CBM extends the CBM architecture into a hierarchical framework. The upper concept and label spaces capture general semantic features and broader class categories, while the lower spaces focus on more specific features and classes.}
\label{fig5}
\end{figure}

Basic-level categories also provide an intuitive entry point for explanation and reasoning in the context of Explainable AI. For instance, it is more natural to explain a prediction as: \textit{``This is a bird with a soft gray back, so it is likely a chickadee''} rather than \textit{``It is likely a chickadee because it has a soft gray back''}. The former structure aligns with human reasoning, as objects are typically first perceived and recognized at the basic-level category~\cite{rosch1976basic}.
Moreover, basic level categories provide multiple advantages, for example, Basic-level categories provide richer and more informative descriptions \cite{markman1997similar}; Members of basic level generally share a common overall shape \cite{rosch1976basic}.

One of the most prominent frameworks for improving interpretability in deep learning is the Concept Bottleneck Model (CBM)~\cite{koh2020concept}. CBMs work by mapping input images to a set of human-understandable concepts through a bottleneck layer, followed by a final classifier that makes predictions based on those concepts. This structure allows users to understand decisions in terms of concept activations. However, existing CBMs typically operate at a single level of semantic abstraction and fail to incorporate hierarchical reasoning, limiting their ability to reflect how humans naturally conceptualize and communicate categories. Recent extensions such as Coarse-to-Fine CBMs (CF-CBM)~\cite{panousis2024coarse} and Spatially-Aware Label-Free CBMs (SALF-CBM)~\cite{benou2025show} introduce mechanisms to incorporate global and local features. They build a hierarchy of visual scale, from global features to local, region-specific features. Yet, they still operate within a flat label space and lack mechanisms to provide concept explanations across multiple abstraction levels.  As a result, these models fall short of enabling full hierarchical reasoning in either the concept or label space. This limitation is especially critical for non-expert users. For instance, a subordinate-level prediction like \textit{snowy egret} is more intuitive if it first references a basic-level category like \textit{bird}. Hierarchical reasoning also supports counterfactual queries within the same superclass (e.g., \textit{``What kind of dog would a golden retriever be if its fur were black?''}), which are important for interpretability.

In this work, we propose a Hierarchical Concept Bottleneck Model with a hierarchy of abstraction. Compared to traditional CBMs, our model provides hierarchical predictions in the label space and generates hierarchical feature based explanations. Importantly, it achieves this without requiring any additional explicit relational labels between high-level and low-level features. Instead, hierarchical information is propagated between the two levels through consistency loss functions that enforce both visual and semantic consistency. The higher-level concept layer acts as a ``novice", identifying general features to classify objects into basic-level categories. In contrast, the lower-level concept layer functions as an ``expert", detecting more specific features to classify objects into subordinate categories within the corresponding basic-level class. In the label space, the model predicts both general categories and specific sub-categories based on the aligned level of abstraction of concepts.
Accordingly, we name our method \textbf{Hierarchical Interpretable Label-Free CBM (HIL-CBM)}: Our proposed HIL-CBM incorporates both a hierarchical concept space and a hierarchical class space. The main contributions of our work are as follows:
\begin{enumerate}
    \item \textbf{Novel framework:} We introduce a new Concept Bottleneck Model architecture that supports hierarchical predictions and explanations across multiple levels of abstraction in both the concept and label spaces.
    
    \item \textbf{Classification performance:} HIL-CBM outperforms existing state-of-the-art CBMs on multiple classification benchmarks -- without requiring explicit relational labels -- by leveraging visual and semantic consistency objectives.

    \item \textbf{Human centric view of interpretability:} The hierarchical structure enables HIL-CBM to align the level of abstraction in concept-based explanations with the model’s predictions. Human evaluation experiments show that HIL-CBM provides greater interpretability and more accurate explanations compared to state-of-the-art CBM.
\end{enumerate}

\section{Related Work}
\label{sec:related_works}

\subsection{Concept Bottleneck Model}
The main goal of interpretability research is to improve human understanding of model decisions and increase trust in intelligent systems \cite{gu2020vinet}. Concept Bottleneck Models (CBMs) are an influential framework for explaining model decisions.
Compared to Standard deep learning classifiers which produce predictions without offering human-understandable justifications,  Concept Bottleneck Model ~\cite{koh2020concept} provides interpretability through a set of human-readable intermediate concepts. CBMs take an image as input, project internal feature representations onto a set of interpretable concepts via a learned bottleneck layer, and make final predictions based solely on these concept activations. This architecture enables users to trace predictions back to semantically meaningful features. For example, users can understand why the model predicts a specific bird species based on the presence of visual characteristics associated with that species.

Despite their interpretability, early CBMs face several limitations. They must be trained from scratch, which is computationally expensive, and depend on the availability of annotated concept labels. To mitigate the retraining burden, Post-hoc CBM~\cite{yuksekgonul2022post} is trained only on the final classifier and an optional residual fitting layer, leveraging Concept Activation Vectors (CAVs)~\cite{kim2018interpretability} and CLIP~\cite{radford2021learning} to extract concept directions from pre-trained image encoders. However, this approach assumes access to the CLIP encoder and is mostly applicable to datasets without concept annotations.

To address the annotation limitation, the Label-Free Concept Bottleneck Model (LF-CBM)~\cite{oikarinen2023label} was introduced. LF-CBM uses GPT-3~\cite{brown2020language} to generate class-level concept descriptions and CLIP-Dissect~\cite{oikarinen2022clip} to train concept layers that align image features with textual concepts. While this removes the need for labeled concept data, LF-CBM still operates at a single level of semantic abstraction.

The Coarse-to-Fine Concept Bottleneck Model (CF-CBM)~\cite{panousis2024coarse} introduces a two level framework that captures features at both whole-image and patch-level. However, its hierarchy is defined as concepts hierarchy and image regions rather than the hierarchy in the class label space. Both levels still support the same downstream flat classification task, rather than predicting labels at different semantic levels. Thus, while CF-CBM introduces hierarchical concept discovery on whole-image and patch-level, it does not introduce a semantic hierarchy on the classes labels.

More recently, the Spatially-Aware Label-Free Concept Bottleneck Model (SELF-CBM)~\cite{benou2025show} enhances interpretability by also predicting the spatial locations where each concept is activated, thereby linking concepts to specific regions in the image. To address the limitation of fixed concept sets, Hybrid-CBM~\cite{liu2025hybrid} combines a static concept bank with a dynamic one, enabling the model to discover new concepts during training and thus expand its conceptual capacity.

\subsection{Prototype Based Explanations for Hierarchical Image Classification}
Hierarchical Image Classification (HIC)~\cite{fan2008mining, yan2015hd,bilal2017convolutional,zhang2022use,novack2023chils,stevens2024bioclip} is a computer vision task in which models classify images according to a hierarchical label structure. To improve the interpretability of hierarchical image classification models, prototype-based HIC models such as~\cite{hase2019interpretable,yu2025hierarchical} learn prototypes at each level of the hierarchy during training and classify images by comparing them to a set of learned, representative prototypes. Hierarchical prototype learning has also been explored in zero shot visual recognition. For example, Zhang et al.~\cite{zhang2019hierarchical} show that structured intermediate prototypes can improve recognition and facilitate knowledge transfer across seen and unseen classes. However, prototype-based models primarily provide visual prototypes for interpretability and do not offer human-readable explanations of model decisions through semantic concepts. In contrast, our proposed HIL-CBM model is able to provide concept-level explanations for model decisions at different levels of the hierarchy in the label space. Since a hierarchical structure also exists in the concept space, the level of abstraction in the feature-based explanations produced by HIL-CBM is aligned with the level of abstraction in the model’s predictions. This alignment enables more intuitive and semantically meaningful interpretations of the model’s decisions. In this work on CBMs, our primary focus is on extending the CBM framework to enhance interpretability, which also highlights a promising pathway for incorporating human-understandable concepts into existing prototype-based HIC models.

\begin{figure*}[t]
\centering
\includegraphics[width=1.0\textwidth]{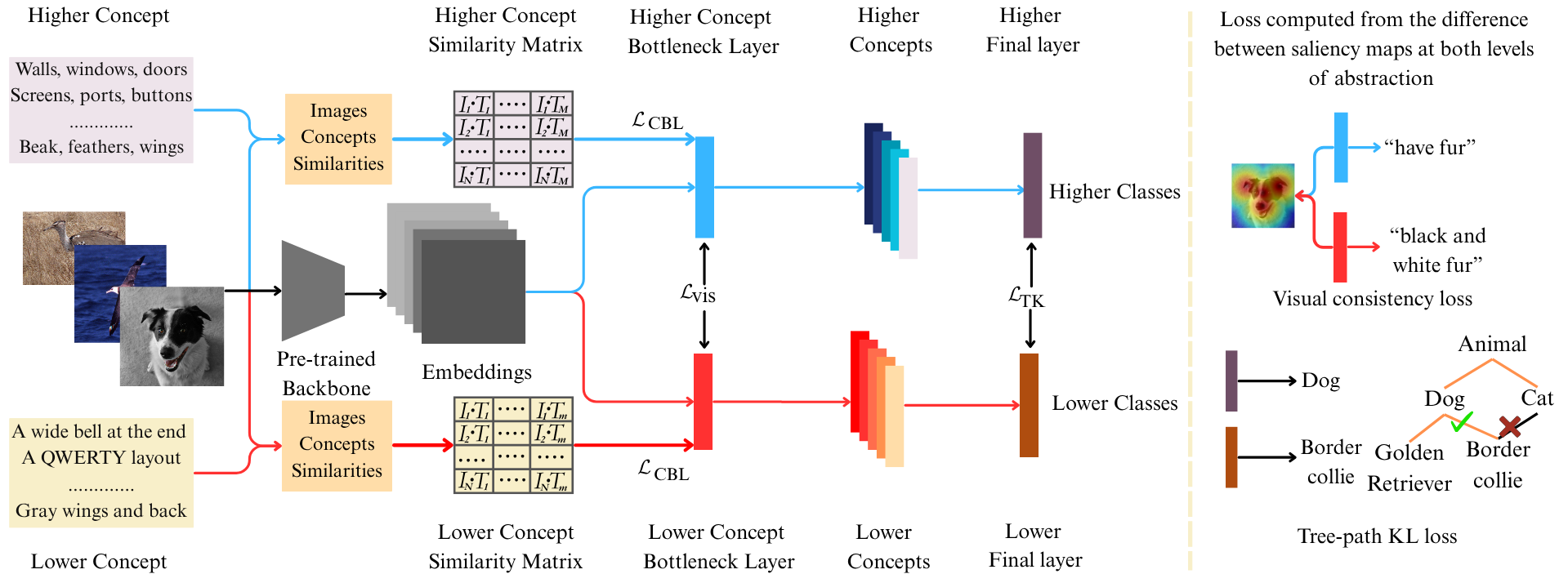} 
\caption{Overview of our proposed model, HIL-CBM. A pre-trained backbone processes input images to produce image embeddings. Two concept layers, each focusing on a different level of abstraction, project the feature maps into interpretable hierarchical concept spaces. These layers are trained using CLIP-Dissect similarity loss and visual consistency loss. Finally, two hierarchical classifiers predict classes at different levels of abstraction, using cross-entropy loss and semantic consistency loss based on the learned hierarchical concept features.}
\label{fig1}
\end{figure*}

\section{Method}
 An overview of our proposed HIL-CBM is shown in Figure~\ref{fig1}. HIL-CBM consists of a pre-trained backbone that transforms input images into embeddings. Two concept layers project these embeddings into two separate concept spaces; the higher-level concept space captures general visual features aligned with broader categories, while the lower-level concept space focuses on specific features aligned with more specific classes. Finally, two sparse classifiers operate on the respective concept spaces to predict labels in a semantically structured hierarchical label space. 

We restrict our hierarchy to two levels: the dataset’s label classes (e.g., Tibetan Mastiff) and their direct parent categories (e.g., Dog). We do not include higher-level abstractions (e.g., Animal), because they rely on vague or non-visual properties (e.g., alive, breathes, moves) that are difficult to ground in image information. Likewise, we exclude intermediate biological taxonomy levels (e.g., Canine), as our goal is to improve the interpretability of CBMs by incorporating hierarchical information, rather than to perform interpretable hierarchical image classification. Additionally, concepts at nearby intermediate levels may be similar, and certain taxonomic names may be too uncommon or unintuitive to enhance user understanding. By focusing on direct parent categories, we capture meaningful semantic structure while avoiding the inclusion of concepts that are either too similar or too vague to provide useful explanations.

\subsection{Generation of Feature Concepts}
To avoid the expensive annotation process, GPT models are commonly used in previous label-free CBM works to construct concept sets \cite{oikarinen2023label,benou2025show}.  Similarly, we use GPT-4~\cite{openai2023gpt4} to generate feature concepts. For datasets that provide only a single level of class labels, we first prompt GPT-4 to group the class labels into higher-level categories using the instruction: ``Group \{class list\} into one level of higher categories".

Feature concepts for both basic-level and subordinate-level classes are obtained as two separate concept sets using the following prompts to GPT-4: ``List the most important features for recognizing something as a \{class\}'' and ``List the things most commonly seen around a \{class\}''. To improve the quality of the concept sets, we follow the filtering method proposed by~\cite{oikarinen2023label}. Specifically, we remove: concepts that are longer than 30 characters, concepts that are overly similar to the original class labels, and concepts that are not highly activated in images in the dataset. 
This results in two separate concept sets: 
$C_L = \{t_{1}^{l}, \dots, t_{n}^{l}\}$ for low-level classes, and 
$C_H = \{t_{1}^{h}, \dots, t_{m}^{h}\}$ for high-level classes. 

\subsection{Hierarchical Concept Bottleneck Layers}
In HIL-CBM, the pre-trained backbone transforms input images into embeddings in a feature space. Two projection layers are then trained to map these features into interpretable concept spaces. The upper projection layer maps the features to concepts corresponding to the higher-level classes, while the lower projection layer maps the same features to concepts corresponding to lower-level classes. This hierarchical structure is inspired by how humans recognize objects: people often identify items at a general level using general features (e.g., \textit{has fur}, \textit{has four legs} for identifying a \textit{dog}). Similarly, the upper concept layer projects the features to general concepts, which are then passed to the higher-level classifier. After this initial recognition, humans may further distinguish subordinate categories using more specific features (e.g., \textit{dog with black and white fur} as a \textit{border collie}). In the same way, the lower concept layer projects the features into concept space, which then predicts the subordinate class.

To ensure that neurons in both concept bottleneck layers are activated by the appropriate concepts, the projection weights are optimized to maximize the similarity between each neuron's activation pattern and the target concept activation. This is implemented using CLIP-based concept supervision: similarities are computed between the projected concept activations and CLIP-Dissect embeddings~\cite{oikarinen2022clip}, which capture the alignment between the CLIP image encoder and the CLIP text encoder~\cite{radford2021learning}. The concept alignment loss for both layers is extended from the cubic cosine similarity loss from \cite{oikarinen2023label} as follows:

\begin{equation}
\begin{aligned}
\mathcal{L}_\text{CBL}
\;=\;
-\sum_{i=1}^{n}
\mathrm{sim}\bigl({\mathbf{q}}_i^{l},\;{\mathbf{P}}_i^{l}\bigr) - \sum_{i=1}^{m}
\mathrm{sim}\bigl({\mathbf{q}}_i^{h},\;{\mathbf{P}}_i^{h}\bigr)
\end{aligned}
\end{equation}
where $\mathrm{sim}(\mathbf{q},\mathbf{P})$ denotes the cubic cosine similarity between the normalized activation vectors of neurons in the concept layer $\mathbf{q}$ and the CLIP-based concept similarity vectors $\mathbf{P}$. $l$ and $h$ refer to the lower and higher abstraction levels, respectively, while $n$ and $m$ represent the number of concepts in the corresponding concept sets at each level.

\subsubsection{Visual consistency}
In the hierarchical structure, the two concept spaces capture different levels of abstraction but are expected to attend to similar regions within the same image (e.g., focusing on the \textit{dog} region when recognizing both \textit{dog} and \textit{border collie}). To enforce visual consistency and promote spatial alignment, we jointly train the two concept layers using shared feature maps from the same pre-trained backbone. Moreover, as saliency maps are widely used in explainable AI to tell the location of the area that model prediction focus on \cite{simonyan2013deep, wang2019learning, peng2023hierarchical, li2023bi, wu2024improving}, we introduce a gradient-based saliency alignment loss to further enhance consistency between the two concept layers without the relational label between the two layers.

Inspired by prior works that use Grad-CAM~\cite{selvaraju2017grad} to align spatial focus with target regions~\cite{selvaraju2019taking, selvaraju2021casting, pillai2022consistent, sheng2025low}, we compute Grad-CAM saliency maps for both concept layers and compare them using mean squared error (MSE) loss. This loss encourages both abstraction levels to focus on overlapping image regions. As such, the gradient-based saliency alignment loss, $\mathcal{L}_{\text{vis}}$ is defined as:

\begin{equation}
\mathcal{L}_{\text{vis}} = 
\left\| \nabla_{f^L_c} \left( \sum_{i=1}^{n} c^{H}_i \right) 
- \nabla_{f^H_c} \left(  \sum_{i=1}^{m} c^{L}_i \right) \right\|_2^2,
\end{equation}
here $c^H$ and $c^L$ denote the concept prediction outputs of the higher-level concept layer $f_{c}^{H}$ and the lower-level concept layer $f_{c}^{L}$, respectively. 

Finally, the loss for the joint concept bottleneck layers training is defined as:

\begin{equation}
\begin{aligned}
\mathcal{L}_{\text{HCBL}} = \mathcal{L}_\text{CBL} +
\lambda_{\text{vis}} \cdot \mathcal{L}_{\text{vis}}
\end{aligned}
\end{equation}
where $\lambda_{\text{vis}}$ denotes the scaling factor of the visual consistency loss. The two concept bottleneck layers, which capture different levels of abstraction, are trained without requiring explicit labels for the relationships between concepts across the two levels.

\subsection{Hierarchical Class Prediction Layer}
Once the two projection layers have been trained, two final layers are trained to predict class labels at both abstraction levels. The upper classifier receives input from the upper concept layer and predicts the higher-level classes. Subsequently, the lower classifier uses the output of the lower concept layer to predict lower-level classes, based on more specific feature representations. The training of the two final classification layers is conducted in two stages. In the first stage, the layers are trained separately to allow each to specialize in its respective level of abstraction. In the second stage, the layers are trained jointly to enhance the consistency between the hierarchical levels.

Since sparsity can improve the interpretability of predictions~\cite{wong2021leveraging}, we aim for the decisions of the model to be explained by a smaller set of feature concepts. Following a similar approach to the training of sparse concept layers in SALF-CBM~\cite{benou2025show} and LF-CBM~\cite{oikarinen2023label}, we train both concept layers separately with two levels of abstraction of feature concepts and class labels using the GLM-SAGA optimizer~\cite{wong2021leveraging}:

\begin{equation}
\begin{aligned}
\mathcal{L} &=\underbrace{\sum_{i=1}^{N} \mathcal{L}_{\text{CE}}\left( \mathbf{W}_F f(\mathbf{x}_i) + \mathbf{b}_F, \; y_i \right)}_{\text{Cross-Entropy Loss}} + \underbrace{\lambda \, \mathcal{R}(\mathbf{W}_F)}_{\text{Elastic net regularization}}
\end{aligned}
\end{equation}

where $W_F$ denotes the weights in the final layer $f$, $\mathbf{b}_F$ is the bias term, $y_i$ is the ground truth label for each level of abstraction, and $\lambda$ is the regularization factor. The elastic net regularization term $\mathcal{R}$ is defined as:

\begin{equation}
\begin{aligned}
\mathcal{R}_\alpha(\mathbf{W}_F) = (1 - \alpha) \cdot \tfrac{1}{2} \| \mathbf{W}_F \|_F^2 + \alpha \| \mathbf{W}_F \|_{1,1}
\end{aligned}
\end{equation}
where $\| \mathbf{W} \|_F^2$ denotes the Frobenius norm, $\| \mathbf{W} \|_{1,1}$ denotes the element-wise matrix norm, and $\alpha$ denotes the blending coefficient used to balance the two components of the regularization term.

\subsubsection{Semantic consistency}
To improve semantic consistency across hierarchical levels and avoid mismatched predictions from the two classifiers (e.g., predicting \textit{golden retriever} at the lower level and \textit{cat} at the higher level), we adopt the Tree-path KL Divergence loss introduced by~\cite{parkvisually}. This loss directly penalizes predictions that are inconsistent with the known class taxonomy, thereby enforcing structural alignment between the predictions of the model and the hierarchical label structure. It encourages the outputs of multiple hierarchical classifiers to follow a valid tree path within the class hierarchy.

We first construct a ground-truth tree-path distribution $\mathbf{Y}$ by concatenating the one-hot encodings of the labels in the both levels: $\mathbf{Y} = \left[\, {\mathbf{Y}}^{L}\; ; {\mathbf{Y}}^{H} \,\right]$ where ${\mathbf{Y}}^{L}$ and ${\mathbf{Y}}^{H}$ denote the one-hot encoding labels at the two levels.

Next, we concatenate the logits output from each classifier and apply the logSoftmax function:

\begin{equation}
\begin{aligned}
\hat{\mathbf{Y}} = \text{LogSoftmax}\left(\left[\, f^{L}\big({c^{\text{L}}}\big)\; ; f^{H}\big({c^{\text{H}}}\big) \,\right]\right)
\end{aligned}
\end{equation}
where $f^{(L)}$ and $f^{(H)}$ denote the classifiers for both levels and $c^{\text{L}}$ and $c^{\text{H}}$ are the features input to that classifiers.

The Tree-path KL Divergence loss is then defined as:

\begin{equation}
\begin{aligned}
\mathcal{L}_{\text{TK}} = \text{KL}(\mathbf{Y} \,\|\, \hat{\mathbf{Y}})
\end{aligned}
\end{equation}

To provide more consistent hierarchical predictions and improved concept-level explanations, we jointly train the two classifiers in the second stage using the Tree-path KL Divergence loss. This objective encourages semantic alignment between the predictions at the higher and subordinate levels by enforcing that they follow a valid path in the class taxonomy. To preserve the sparsity induced during the first training stage, we mask the zero weights in both classification layers, restricting updates during this phase to only the non-zero weights from the sparse concept layers. This prevents the reintroduction of noisy or non-contributing features whose connections were previously suppressed, thereby maintaining interpretability through a compact and semantically meaningful concept representation. 

The loss function is defined as:

\begin{equation}
\begin{aligned}
\mathcal{L}_{\text{total}} = \mathcal{L}^\text{H}_{\text{CE}} + \mathcal{L}^\text{L}_{\text{CE}} + \lambda_{semantic} \mathcal{L}_{\text{TK}} 
\end{aligned}
\end{equation}

where $\mathcal{L}^\text{H}_{\text{CE}}$ is the cross-entropy loss for the higher-level classifier, which predicts the basic-level classes, and $\mathcal{L}^\text{L}_{\text{CE}}$ is the cross-entropy loss for the lower-level classifier, which predicts the subordinate-level classes. $\lambda_{semantic}$ is the scaling factor for the semantic consistency loss $\mathcal{L}_{\text{TK}}$.
\section{Experiments \& Results}
In this section, we present the evaluation results of HIL-CBM to demonstrate the model’s performance in terms of both classification accuracy and interpretability. We then evaluate the visual and semantic consistency between the two hierarchical layers. Finally, a human evaluation is conducted to assess whether the hierarchical structure enhances the interpretability of the Concept Bottleneck Model.

\subsection{Datasets}
We train HIL-CBM on four benchmark datasets that vary in size, resolution, and number of classes. CIFAR-100~\cite{krizhevsky2009learning} includes 100 object classes with approximately 50,000 low-resolution training images. CUB-200~\cite{wah2011caltech} consists of 200 fine-grained bird species with about 5,900 training samples. Places365~\cite{zhou2017places} is a large-scale scene classification dataset comprising around 1.8 million images of both indoor and outdoor environments. Finally, ImageNet~\cite{deng2009imagenet} contains 1,000 object categories across diverse domains, with approximately 1.2 million training images. 

\subsection{Setup}

\begin{table*}[h!]
\centering
\caption{The performance of our proposed model, HIL-CBM, compare with state-of-the-art Concept Bottleneck Models.
}
\begin{tabular}{lcccccc}
\toprule
\textbf{Model} & \textbf{Backbone} & \textbf{Sparse} & \multicolumn{4}{c}{\textbf{Lower Level$\Vert$\ Higher Level} } \\
\cmidrule(lr){4-7} &&& \textbf{CIFAR100} & \textbf{CUB-200} & \textbf{Places365} & \textbf{ImageNet} \\
\midrule
Standard & ResNet & Yes & 58.34\% & \textbf{75.96\%} & 38.46\% & 74.35\% \\
Standard Hierarchical & ResNet & Yes & 58.60\% $\Vert$\ 72.39\% & 75.92\%\ $\Vert$\ 83.78\% & 40.42\%\ $\Vert$\ 62.66\% & 74.36\% $\Vert$\ 82.46\% \\
P-CBM & ResNet & Yes & 43.20\% & 59.60\% & -- & -- \\
LF-CBM & ResNet & Yes & 65.13\% & 74.31\% & 43.68\% & 71.95\% \\
HybridCBM & ResNet & Yes & 68.38\% & -- & -- & -- \\
SALF-CBM & ResNet & Yes & -- & 74.35\% & 46.73\% & 75.32\% \\
HIL-CBM (Ours) & ResNet & Yes & \textbf{69.50\%} $\Vert$\ 73.55\% & 75.35\% $\Vert$\ 83.08\% & \textbf{47.59\%} $\Vert$\ 57.91\%
 & \textbf{75.54\%} $\Vert$\ 81.50\%\\ 
\midrule
CF-CBM(Image\&Class names)& ViT & Yes & -- & 79.50\% & --& 77.40\%\\
HIL-CBM(Class names)& ViT & Yes & 79.24\%$\Vert$\ 86.04\%  &\textbf{79.55\%}$\Vert$\ 82.95\% & 53.95\% $\Vert$\ 69.61\% &\textbf{78.71\%}$\Vert$\ 84.60\%\\
\midrule
CF-CBM(Patches\&Features)& ViT & Yes & -- & 73.20\% & -- &78.45\%\\
HIL-CBM(Features)& ViT & Yes & 81.54\% $\Vert$\ 86.12\% &\textbf{75.34\%} $\Vert$\ 86.96\% & 54.99\% $\Vert$\ 69.71\% &\textbf{79.43\%} $\Vert$\  84.64\%\\
\midrule
Standard & ResNet & No & 70.10\% & 76.70\% & 48.56\% & 76.13\% \\
SALF-CBM & ResNet & No & -- & 76.21\% & 49.38\% & 76.26\% \\
HIL-CBM(Ours) & ResNet & No & 69.91\% $\Vert$\ 72.36\% & 75.44\% $\Vert$\ 82.26\% & 48.12\% $\Vert$\ 57.54\% & 75.18\% $\Vert$\ 81.04\% \\
\bottomrule
\end{tabular}
\label{tab:ACC}
\end{table*}

To allow our model to be evaluated under the same conditions as prior works~\cite{oikarinen2023label,benou2025show}, we use the same backbone architectures as those employed by existing state-of-the-art models for each benchmark dataset. Specifically, for training on the CIFAR-100 dataset, we adopt the CLIP image encoder with the RN50 variant as the backbone \cite{radford2021learning}, while for ImageNet and Places365, we use a standard ResNet-50~\cite{he2016deep} backbone. Also, for fair comparison, we adopt the concept sets provided by~\cite{oikarinen2023label} as the candidate subordinate-level concepts for CIFAR-100, CUB-200, Places365, and ImageNet. To further compare our results with existing CBMs that use a ViT-based backbone, we also train our model on the four datasets using the CLIP image encoder with the ViT-B/16 variant as the backbone.

For CIFAR-100, we use the 20 superclass labels provided in the dataset as the higher-level categories. For datasets such as ImageNet, which contain general images spanning a wide range of categories, we use GPT-4~\cite{openai2023gpt4} to group the fine-grained classes into broader, basic-level categories. Similarly, for domain-specific datasets like CUB-200 and Places365, which focus on bird species and scene locations, respectively, we employ GPT-4 to organize the classes into higher-level categories based on bird taxonomy and semantic scene groupings. At the higher level of abstraction, the 1,000 classes in ImageNet are grouped into 61 basic-level categories; the 200 classes in CUB-200 into 70 categories; and the 365 classes in Places365 into nine categories. The corresponding higher-level concept sets include 161 concepts for CIFAR-100, 559 concepts for CUB-200, 99 concepts for Places365, and 366 concepts for ImageNet.

\subsection{Quantitative Evaluations}
Table~\ref{tab:ACC} presents the classification accuracy of our proposed model on the four benchmark datasets. The results for our model are reported as the mean values over three training runs. To ensure a fair comparison, we train HIL-CBM under different training settings to compare it with existing CBMs using the same backbone and sparse final-layer setting. We compare our results against the following baselines: (1) the standard backbone model, with and without a sparse final layer for class label prediction (results reported from \cite{oikarinen2023label});
(2) the standard backbone model, with two sparse hierarchical classifiers;
(3) Post-hoc CBM (P-CBM)~\cite{yuksekgonul2022post};
(4) Label-Free CBM (LF-CBM)~\cite{oikarinen2023label};
(5) Coarse-to-Fine CBM (CF-CBM)~\cite{panousis2024coarse};
(6) Hybrid CBM (Hybrid-CBM)~\cite{liu2025hybrid};
(7) Spatially-Aware Label-Free CBM (SALF-CBM)~\cite{benou2025show}.

Our proposed model, HIL-CBM, outperforms all state-of-the-art CBMs under the sparse final layer setting. With the ResNet based backbone, on the CIFAR-100, Places365, and ImageNet datasets, HIL-CBM demonstrates superior performance compared to the standard backbone model with a sparse classification layer. On the CUB-200 dataset, HIL-CBM achieves comparable accuracy to the sparse backbone model. These results indicate that HIL-CBM is capable of delivering hierarchical, interpretable explanations without compromising classification accuracy. Furthermore, when comparing the performance of HIL-CBM with a sparse versus a dense final layer across different datasets, we observe only a small performance difference. This suggests that our model can effectively learn meaningful and sufficient explanations for the target classes in the sparse setting. With the ViT-based backbone, we compare HIL-CBM against existing CBM models trained with the same backbone. HIL-CBM also outperforms state-of-the-art CBMs across multiple training settings.

\subsection{Ablation Study}
Our proposed model incorporates a visual consistency loss to enforce alignment between the two levels of abstraction in the concept space during the joint training of the concept layers. Without requiring explicit relational labels between the two levels, this loss encourages both concept layers to focus on similar regions of the input images, thereby improving consistency. In the label space, a semantic consistency loss is introduced to promote coherence between the predictions at the basic and subordinate levels.

To evaluate the effects of the consistency losses and the joint training process, we conduct several experiments for ablation studies. (1) We train the model without the visual consistency loss, so that the two concept layers are optimized solely using the cubic-cosine similarity loss at each level. We also train the model without the semantic consistency loss, in which case the two classifiers are trained independently without the joint training stage. (2) We train the model with different combinations of the visual and semantic consistency loss weights to identify an effective setting for the two loss terms. (3) We train the model with different types of visual consistency losses to examine the effect of the specific alignment objective. (4) We train the model without the semantic consistency loss to evaluate the contribution of the joint training process. (5) We train existing CBMs with a single flat concept space and label space to determine whether the observed accuracy gain comes from the hierarchical structure itself or simply from the increased concept budget.

\subsubsection{Consistency Loss}
Table~\ref{tab:acc2} presents the accuracies of the higher- and lower-level classifiers on CIFAR-100, CUB-200, Places365, and ImageNet under four training settings: \textit{Neither}, with no joint training and neither consistency loss; \textit{Visual only}, with only the visual consistency loss; \textit{Semantic only}, with only the semantic consistency loss; and \textit{Both}, with joint training and both consistency losses. The results demonstrate the benefit of incorporating alignment across the two abstraction levels. In particular, both consistency losses improve prediction accuracy at both levels, even without cross-level masking.

\begin{table}[h!]
\centering
\caption{Higher and lower-level classifiers’ accuracy under four training settings.
}
\begin{tabular}{ccccccc}
\toprule
\textbf{With}  &  &  \textbf{CIFAR-100} &\textbf{CUB}&\textbf{Place365}&\textbf{ImageNet} \\
\midrule
\multirow{2}{*}{\text{Neither}}&Higher &73.45 & 81.35 &57.45    & 75.55  \\
&Lower &68.83  & 73.90 &45.41  & 72.03\\
\midrule
\multirow{2}{*}{{\text{Visual}}} &Higher &\textbf{73.65}  &82.47  &57.32 &77.62 \\
& Lower & 69.23&74.80& 44.09&72.74  \\
\midrule
\multirow{2}{*}{\text{Semantic}}&Higher &73.31   & 82.75&57.37   & 78.43  \\
&Lower  &69.12 & 74.75 &46.35 & 74.90  \\
\midrule
\multirow{2}{*}{{\text{Both}}}&Higher &73.55  & \textbf{83.08}&\textbf{57.91}   & \textbf{81.50}  \\
&Lower &\textbf{69.50}  & \textbf{75.35} &\textbf{47.59}  & \textbf{75.54}  \\
\bottomrule

\end{tabular}
\label{tab:acc2}
\end{table}

\subsubsection{Consistency Loss Weights}
To determine the two consistency loss weights, we train our proposed model with different combinations of visual and semantic weights on the two benchmark datasets most commonly used by existing CBMs: CUB-200 and ImageNet. Table~\ref{tab:weights} shows the ablation results for these settings. We report classification accuracy on both datasets for various weight combinations. The results show that a semantic weight of 0.1 and a visual weight of 0.7 yield the best performance on both datasets.

\begin{table*}[t]
\centering
\caption{Results under different visual and semantic weights on CUB-200 and ImageNet datasets.}
\begin{tabular}{c|ccc|ccc}
\hline
& \multicolumn{3}{c|}{\textbf{CUB-200}} & \multicolumn{3}{c}{\textbf{ImageNet}} \\
\textbf{Visual/Semantic} & Semantic 0.1 & Semantic 0.3 & Semantic 0.5 
& Semantic 0.1 & Semantic 0.3 & Semantic 0.5 \\
\hline
Visual 0.1 & 75.27\% & 74.92\% & 74.82\% & 75.36\% & 75.26\% & 75.17\% \\
Visual 0.3 & 75.01\% & 75.03\% & 75.08\% & 75.39\% & 75.37\% & 75.30\% \\
Visual 0.5 & 74.78\% & 74.91\% & 75.01\% & 75.48\% & 75.40\% & 75.32\% \\
Visual 0.7 & \textbf{75.35\%} & 74.92\% & 75.03\% & \textbf{75.54\%} & 75.35\% & 75.29\% \\
\hline
\end{tabular}

\label{tab:weights}
\end{table*}

\begin{table}[t]
\centering
\caption{Results of HIL-CBM with two different visual consistency loss settings}
\begin{tabular}{c|ccc|ccc}
\hline
\textbf{Datasets} & {\textbf{CUB-200}} & {\textbf{ImageNet}} \\

\hline
IoU Based & 74.77\% $\Vert$\ 82.95\%& 73.52\% $\Vert$\ 80.93\% \\
MSE Based & 75.53\% $\Vert$\ 83.08\% & 75.54\% $\Vert$\ 81.50\% \\
\hline
\end{tabular}

\label{tab:visual}
\end{table}
\subsubsection{Visual Consistency} To avoid expensive relational annotations between features at different levels of abstraction, we introduce a Grad-CAM-based alignment loss in our proposed model. This loss encourages both abstraction levels to attend to overlapping image regions. To evaluate the effect of visual consistency, we trained HIL-CBM with two different visual consistency losses. Table~\ref{tab:visual} represents the results with the two different training setting on CUB-200 and ImageNet. With the IoU-based consistency loss, the model achieved 74.77\% lower-level accuracy and 82.95\% higher-level accuracy on CUB-200, and 73.52\% lower-level accuracy and 80.93\% higher-level accuracy on ImageNet. For both datasets, these results were lower than those obtained with the MSE-based consistency loss.

\subsubsection{Semantic Consistency} Without the strong masking from higher to lower levels, we train the model with a Tree-path loss which encourage the two level prediction follow a solid path inside the taxonomies to enhance the consistency between the two levels predictions. To evaluate the hierarchical consistency gain of the semantic loss, we evaluated hierarchical consistency under two settings: the full model and without the semantic consistency training. We evaluate hierarchical consistency using two complementary metrics based on the relationship between fine-level and coarse-level predictions. The consistency is evaluated with two consistency metrics.
\begin{itemize}
\item \textbf{Model consistency:} It measures whether the lower-level prediction maps to the same higher-level category as the model’s predicted higher-level label. It reflects the agreement between the two classification heads.
\item \textbf{Ground-truth consistency:} It measures whether the lower-level prediction maps to the ground-truth higher-level label. It reflects the semantic correctness of the lower-level prediction with respect to the hierarchy.
\end{itemize}
Table~\ref{tab:semantic} presents the results of the model evaluated under the two consistency metrics. On CUB, 89.64\% of low-level predictions matched the model’s high-level category, compared to 87.51\% without the loss; on ImageNet, consistency improved from 80.75\% to 84.40\%. When comparing low-level predictions to the ground-truth higher-level labels, CUB achieved 84.32\% consistency with the loss versus 83.96\% without it, and ImageNet achieved 87.77\% versus 86.17\%. The results demonstrate that joint training improves consistency between the two levels of abstraction without requiring relational masking.

\begin{table}[t]
\centering
\caption{Effect of semantic consistency loss on hierarchical consistency.}
\label{tab:semantic}
\begin{tabular}{l l c c}
\hline
\textbf{Metric} & \textbf{Semantic} & \textbf{CUB-200} & \textbf{ImageNet} \\
\hline
\multirow{2}{*}{\text{Model consistency}} & Without & 87.51\% & 80.75\% \\
 & With  & \textbf{89.64\%} & \textbf{84.40\%} \\
\hline
\multirow{2}{*}{\text{Ground-truth consistency}} & Without  & 83.96\% & 86.17\% \\
 & With  & \textbf{84.32\%} & \textbf{87.77\%} \\
\hline
\end{tabular}
\end{table}

\subsubsection{Concept Budget} Since our proposed HIL-CBM projects image features into hierarchical concept spaces at different levels of abstraction, one may ask whether the observed accuracy improvement comes from the hierarchical structure itself or simply from using a larger number of concepts. To investigate this, we trained an existing CBM with a single flat concept space and label space while increasing its concept budget to match the total number of concepts used across both levels of HIL-CBM.
Specifically, we trained LF-CBM~\cite{oikarinen2023label} on CUB-200 and ImageNet using the same total number of concepts as in our two-level model. With the enlarged concept set, LF-CBM improved only marginally, from 74.31\% to 74.39\% on CUB-200 and from 71.95\% to 72.09\% on ImageNet. These results suggest that simply increasing the number of concepts in a flat concept space provides only limited benefit. This comparison indicates that the performance gain of HIL-CBM is not merely due to a larger concept budget, but rather to the hierarchical structure itself. Moreover, both LF-CBM and our proposed model use sparse final classifiers for label prediction based on concepts. Because the sparse classifier relies on only a small subset of concepts for each class, increasing the total number of available concepts yields only slight improvements in accuracy.

\subsection{Interpretability Validation}

The most important characteristic of our HIL-CBM is its hierarchical structure in both the feature space and the label space. The model is capable of predicting labels at two levels of abstraction while simultaneously providing feature based explanations at the corresponding levels of abstract concept spaces. To quantitatively evaluate the benefits of this hierarchical structure in terms of interpretability, we conduct a user study comparing HIL-CBM to models with a single-level structure. Our work focuses on introducing hierarchy in both the concept space and the label space to enhance the interpretability of CBMs. To avoid the costly process of feature and relational label annotation, we keep HIL-CBM in a label-free setting. The main idea behind HIL-CBM is the utility of hierarchical structure in providing feature based explanations at different levels of abstraction, corresponding to label predictions at the same levels of abstraction. For a fair evaluation of how this design enhances the interpretability of CBMs, we choose LF-CBM \cite{oikarinen2023label}, since it operates in a fully label-free setting and provides text-based explanations. By contrast, CF-CBM~\cite{panousis2024coarse} is trained on annotated concepts, and its structure relies on class-concept masking supervision. Similarly, SALF-CBM~\cite{benou2025show} provides heatmap based explanations in addition to text, whereas our model focuses only on the text modality, making the it less directly comparable.

In this experiment, we randomly select 35 images from the ImageNet validation set and use them as input to our model. We compare the feature concept explanations generated by our proposed HIL-CBM model with those produced by the baseline LF-CBM model~\cite{oikarinen2023label}. A total of 50 human raters from Prolific are asked to evaluate the outputs of both models along two dimensions, using a 5-point Likert scale from 1 to 5. The evaluation criteria are as follows:
\begin{enumerate*}[label=(\Alph*)]
    \item To what extent does the model’s explanation help you understand why it classified the image as the \textit{ground truth class}?  
    \item To what extent do you agree that the model provides an accurate explanation?
\end{enumerate*}

Table~\ref{tab:likert-comparison} presents the results of the user study comparing our proposed HIL-CBM with the baseline LF-CBM model. For both evaluation criteria, HIL-CBM achieved higher average scores compared to the baseline. These results demonstrate that HIL-CBM not only offers improved decision interpretability but also provides more accurate and semantically meaningful concept based explanations. 

\begin{table}[t]
\centering
\caption{Comparison of Likert-scale responses between LF-CBM and HIL-CBM for criteria A and B.}
\begin{tabular}{llcc}
\cmidrule(l){1-4}
 &  & \multicolumn{2}{c}{\textbf{Criterion}} \\
\textbf{Score} & \textbf{Model} & \textbf{A} & \textbf{B} \\ \cmidrule(l){1-4}
Mean $\pm$ std & LF-CBM & \begin{tabular}[c]{@{}c@{}}2.85\\ $\pm$1.28\end{tabular} & \begin{tabular}[c]{@{}c@{}}2.84\\ $\pm$1.2\end{tabular} \\ \cmidrule(l){2-4}  
 & HIL-CBM & \begin{tabular}[c]{@{}c@{}}3.39\\ $\pm$1.22\end{tabular} & \begin{tabular}[c]{@{}c@{}}3.38\\ $\pm$1.20\end{tabular} \\ \cmidrule(l){1-4}
\multicolumn{4}{l}{\textbf{Response Distribution (\%)}} \\ \cmidrule(l){1-4}
1. None at All & LF-CBM & 19.09 & 17.09 \\
 & HIL-CBM & 9.66 & 8.23 \\ \cmidrule(l){2-4}
2. A Little & LF-CBM & 21.94 & 26.00 \\
 & HIL-CBM & 12.80 & 15.14 \\ \cmidrule(l){2-4} 
3. A Moderate Amount & LF-CBM & 25.37 & 24.06 \\
 & HIL-CBM & 27.03 & 27.66 \\ \cmidrule(l){2-4}  
4. A Lot & LF-CBM & 22.17 & 21.89 \\
 & HIL-CBM & 29.66 & 28.80 \\ \cmidrule(l){2-4} 
5. A Great Deal & LF-CBM & 11.43 & 10.97 \\
 & HIL-CBM & 20.86 & 20.17 \\ \cmidrule(l){1-4}
\end{tabular}%
\label{tab:likert-comparison}
\end{table}

\subsection{Hierarchical Explainable Classification}
\begin{figure}[t]
\centering
\includegraphics[width=0.47\textwidth]{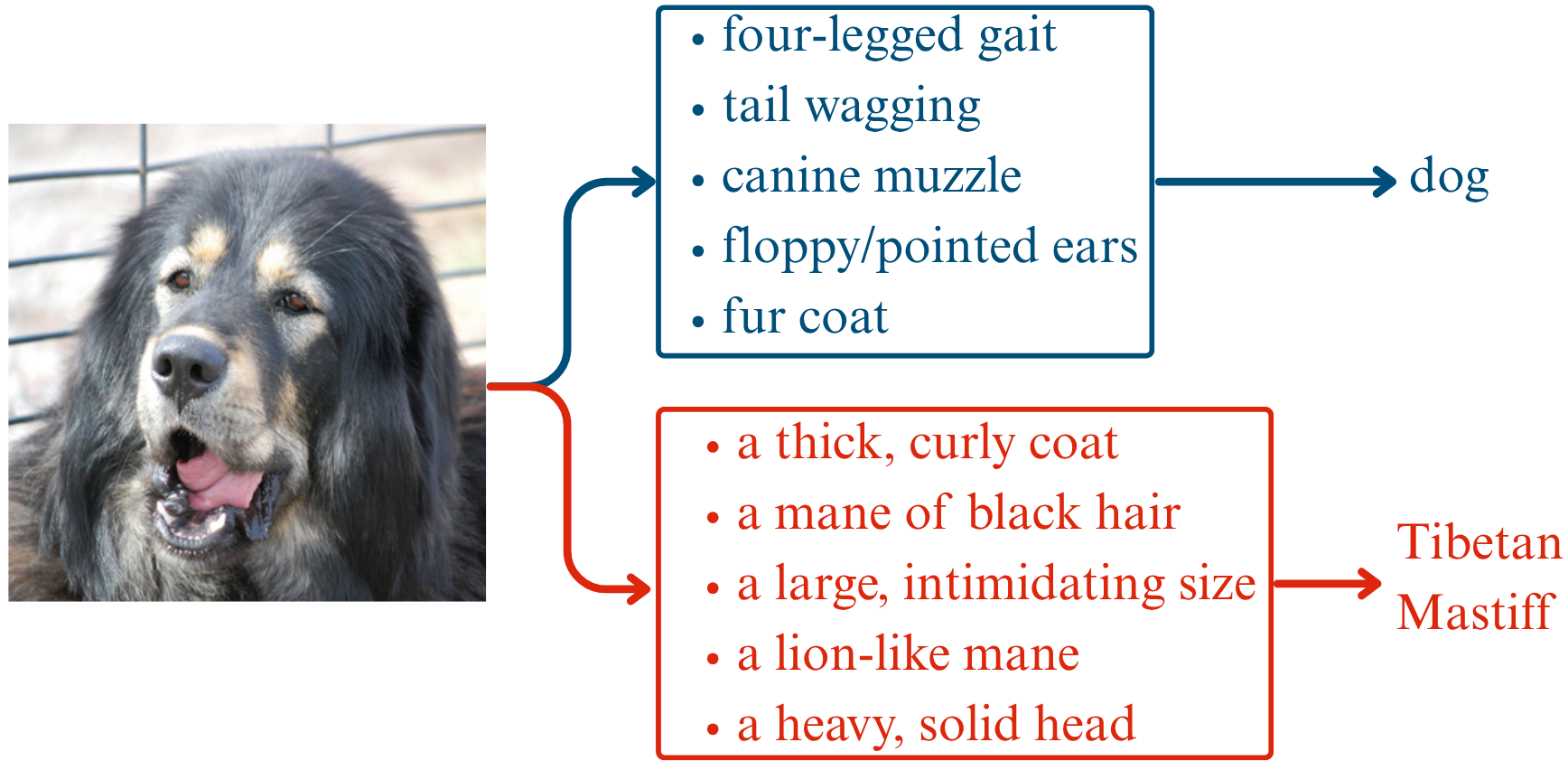} 
\caption{Visualization of two-level predictions from HIL-CBM and the corresponding explanations at each level. The model predicts both levels of class labels, accompanied by concept-based explanations aligned with each level of abstraction. This hierarchical interpretability enables users to understand the model decisions from general to specific.}
\label{fig3}
\end{figure}

To demonstrate the interpretability of our proposed model, we visualize the explanation process for a sample prediction of the class \textit{Tibetan Mastiff} from the ImageNet dataset. Figure~\ref{fig3} illustrates the model’s interpretability across both levels of abstraction. Given the input image, the model predicts class labels at both the basic and subordinate levels. For each prediction, corresponding explanations are provided through two distinct levels of concept-based abstraction, illustrating the hierarchical interpretability enabled by our model. Additional visualizations are provided in the supplementary material.

\begin{figure}[t]
\centering
\includegraphics[width=0.47\textwidth]{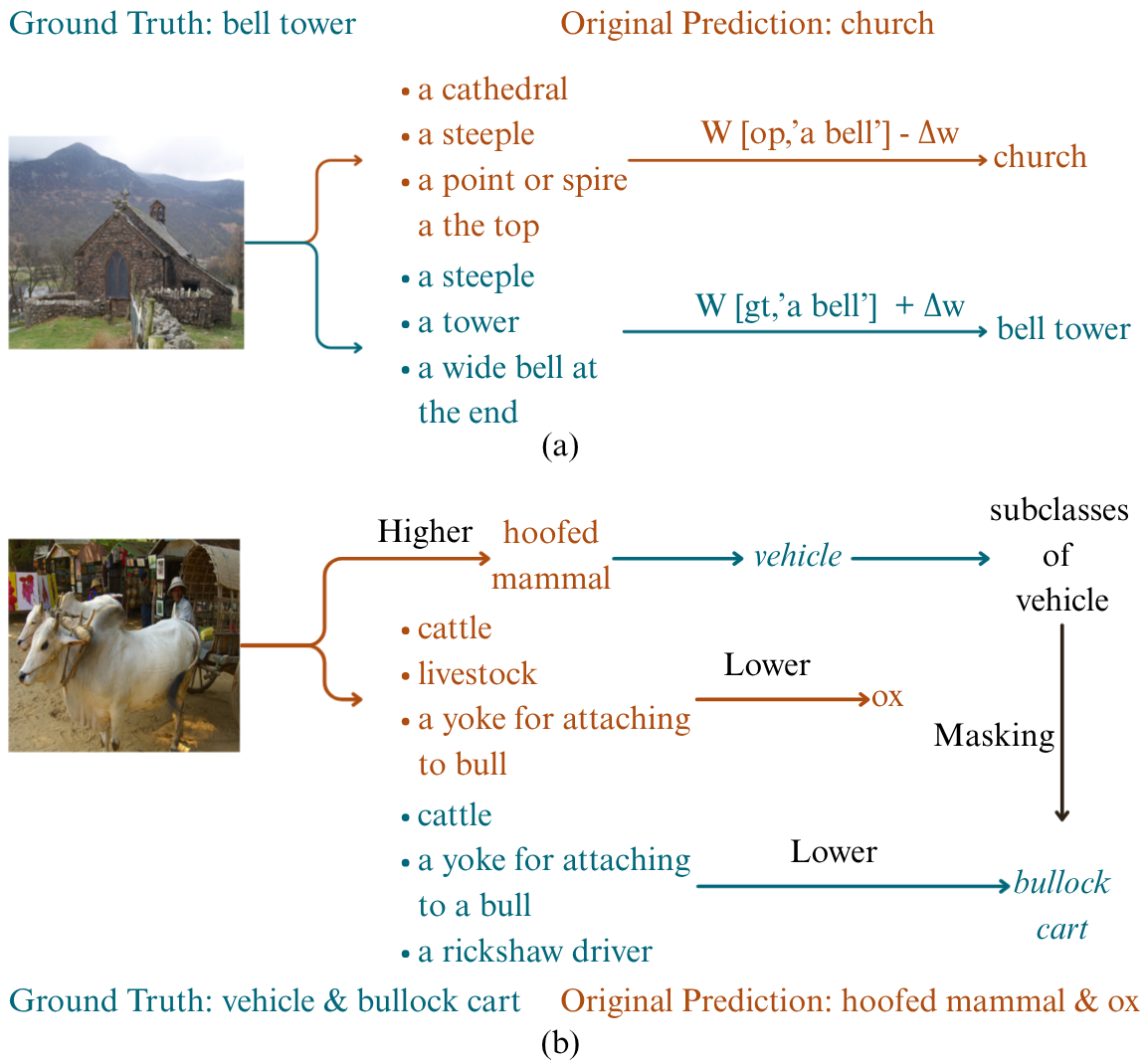} 
\caption{
Examples of model debugging. 
\textbf{(a)} An example where the higher-level prediction is correct, but the lower-level prediction is wrong. Debugging is performed by editing weights using domain knowledge. 
\textbf{(b)} An example where both predictions are incorrect. Debugging proceeds hierarchically: correcting the higher-level prediction enables guided refinement of the lower-level prediction via class masking. 
}

\label{fig4}
\end{figure}
Figure~\ref{fig4} visualizes the model debugging process using an example from the ImageNet validation set. In Figure~\ref{fig4} (a), the higher-level (basic) class prediction is correct, but the subordinate-level prediction is incorrect. Such errors typically occur between visually similar subclasses. To correct the model, domain experts can identify key visual concepts that distinguish the classes. In this case, the concept \textit{a bell} initially leads the model to predict \textit{church} incorrectly. By reducing the weight to \textit{church} and increasing the weight to the correct class \textit{bell tower}, the model is guided toward the correct prediction. With the updated weights, the model can better disambiguate between similar subordinate categories.

In Figure~\ref{fig4} (b), both the basic-level and subordinate-level predictions are incorrect. Debugging is performed hierarchically by first correcting the basic-level prediction. Once the correct higher-level class is identified, a class mask is applied to constrain the subordinate-level classifier to consider only classes within the corrected higher-level category. Without manually editing the lower-level weights, the model can produce a new, correct subordinate-level prediction aligned with the higher-level class, demonstrating how HIL-CBM enables structured and interpretable model debugging.
At the same time, users can also explore counterfactual explanations within the same higher-level category. For example, within the basic-level category \textit{dog}, one might ask: \textit{What would the model predict if a border collie had golden fur?} This form of reasoning is a key aspect of interpretability, as it enables users to understand how specific feature changes influence the model's predictions among similar subordinate classes within the same higher-level category. As noted in~\cite{oikarinen2023label}, selecting concepts for intervention is challenging and often depends heavily on domain expertise. Unlike standard predictive evaluation, where effectiveness can be measured directly with quantitative metrics, the utility of concept intervention for model debugging is inherently difficult to assess in a systematic evaluation. The choice of intervened concepts is often subjective, task dependent, and influenced by the user's knowledge of the domain. Consequently, it is difficult to design quantitative experiments that fairly and comprehensively to evaluate on the concept-based debugging.

For this reason, prior work has generally emphasized qualitative rather than quantitative evaluation in this setting. Under the same consideration, providing a comprehensive and fair comparison of the effect of hierarchical structure on model debugging between our proposed model and existing CBMs is also difficult. The hierarchical design of our model affects not only the explanations themselves but also the way users interact with the model, since it enables inspection and intervention at multiple levels of abstraction. Such flexibility may improve the debugging process by allowing users to trace errors from coarse abstract concepts to finer-grained ones. However, these benefits are difficult to summarize with a single quantitative metric and may vary depending on the task and the user's expertise. Therefore, we regard the evaluation of hierarchical structure for model debugging primarily as a qualitative question, while leaving a more standardized quantitative assessment for future work.

\section{Conclusion}
In this work, we presented HIL-CBM, the Concept Bottleneck Model, to explicitly model different levels of abstraction in both the concept space and the label space. Experiments on benchmark datasets demonstrated that HIL-CBM outperforms existing state-of-the-art sparse CBMs across several classification tasks. The hierarchical structure is learned using a Grad-CAM-based visual consistency loss in the concept space, without requiring explicit concept relationship labels. Additionally, a tree-path semantic consistency loss is employed to enhance alignment between hierarchical levels in the label space.

Finally, we introduced the hierarchical reasoning capability of the model, which supports counterfactual explanations and debugging within the same higher-level category. We believe this ability is particularly beneficial in domains where subordinate classes exhibit high visual similarity. Furthermore, model performance could be improved through expert intervention by manually refining concept weights. In this work, we focus on enhancing the interpretability of CBMs and restrict our hierarchy to two levels to enable the model capture meaningful semantic structure
while avoiding the inclusion of concepts that are either too similar or too vague.In future work, we aim to explore the potential to extend our framework to provide interpretability to Hierarchical image classification.

\bibliographystyle{IEEEtran}
\bibliography{bib}

\end{document}